\documentclass[conference]{IEEEtran}
\IEEEoverridecommandlockouts
\usepackage{cite}
\usepackage{amsmath,amssymb,amsfonts}
\usepackage{algorithmic}
\usepackage{graphicx}
\usepackage{textcomp}
\usepackage{booktabs}
\usepackage{multirow}
\usepackage{subfig}
\usepackage{url}
\usepackage{adjustbox}
\usepackage{xcolor}
\def\BibTeX{{\rm B\kern-.05em{\sc i\kern-.025em b}\kern-.08em
    T\kern-.1667em\lower.7ex\hbox{E}\kern-.125emX}}
\begin{document}

\title{Linguistic-style-aware Neural Networks for \\ Fake News Detection}

\author{Xinyi Zhou\textsuperscript{1}\textsuperscript{*}\textsuperscript{\textdagger}, Jiayu Li\textsuperscript{2}\textsuperscript{*}, Qinzhou Li\textsuperscript{3}\textsuperscript{\textdagger}, Reza Zafarani\textsuperscript{2}\\
\textsuperscript{1}Paul G. Allen School of Computer Science and Engineering, University of Washington, Seattle, WA 98195 \\
\textsuperscript{2}Department of Electrical Engineering and Computer Science, Syracuse University, Syracuse, NY 13244 \\
\textsuperscript{3}Google, Durham, NC 27701 
\thanks{\textsuperscript{*}The authors are equally contributed.}
\thanks{\textsuperscript{\textdagger}This work was completed while the authors were at Syracuse University.}}

\maketitle

\begin{abstract}
We propose the hierarchical recursive neural network (HERO) to predict fake news by learning its linguistic style, which is distinguishable from the truth, as psychological theories reveal. We first generate the hierarchical linguistic tree of news documents; by doing so, we translate each news document's linguistic style into its writer's usage of words and how these words are recursively structured as phrases, sentences, paragraphs, and, ultimately, the document. By integrating the hierarchical linguistic tree with the neural network, the proposed method learns and classifies the representation of news documents by capturing their locally sequential and globally recursive structures that are linguistically meaningful. It is the first work offering the hierarchical linguistic tree and the neural network preserving the tree information to our best knowledge.
Experimental results based on public real-world datasets demonstrate the proposed method's effectiveness, which can outperform state-of-the-art techniques in classifying short and long news documents. We also examine the differential linguistic style of fake news and the truth and observe some patterns of fake news.\footnote{The code and data are available at \url{https://github.com/Code4Graph/HERO}.}
\end{abstract}


\section{Introduction}
\label{sec:introduction}

``Fake news,'' as deceptive and misleading news articles (or statements at times), has been broadly discussed along with its influence on democracies and economies~\cite{zhou2020survey}. Public health has also been negatively impacted, especially with the ``infodemic'' that we face along with the pandemic~\cite{zhou2022fake}. Effective fake news detection has thus become an urgent task to mitigate such detrimental impacts.

Psychological theories, such as \textit{Undeutsch hypothesis}~\cite{undeutsch1967beurteilung}, have suggested that the linguistic style of fake news is distinguishable from that of the truth. Therefore, effective techniques can be designed to identify fake news by analyzing the linguistic style of news articles~\cite{zhou2020survey}. Linguistic style can be captured by looking at the writer's usage of words (lexically and semantically) and the way these words are further formed into sentences (syntactic level) and the document  (discourse level)~\cite{zhou2020survey}. Within a machine learning framework, existing studies have captured a news article's linguistic style by computing the frequencies of each word~\cite{perez2018automatic,zhou2020fake}, part of speech (POS, at the syntactic level)~\cite{potthast2018stylometric,przybyla2020capturing,zhou2020fake}, and rhetorical relationship (RR, at the discourse level)~\cite{rubin2015truth,zhou2020fake}. These frequencies form a news article's representation, which is further classified by, e.g., support vector machines (SVM) and random forests to predict the news as fake news or the truth.

\begin{figure*}[t]
    \centerline{\includegraphics[width=0.9\textwidth]{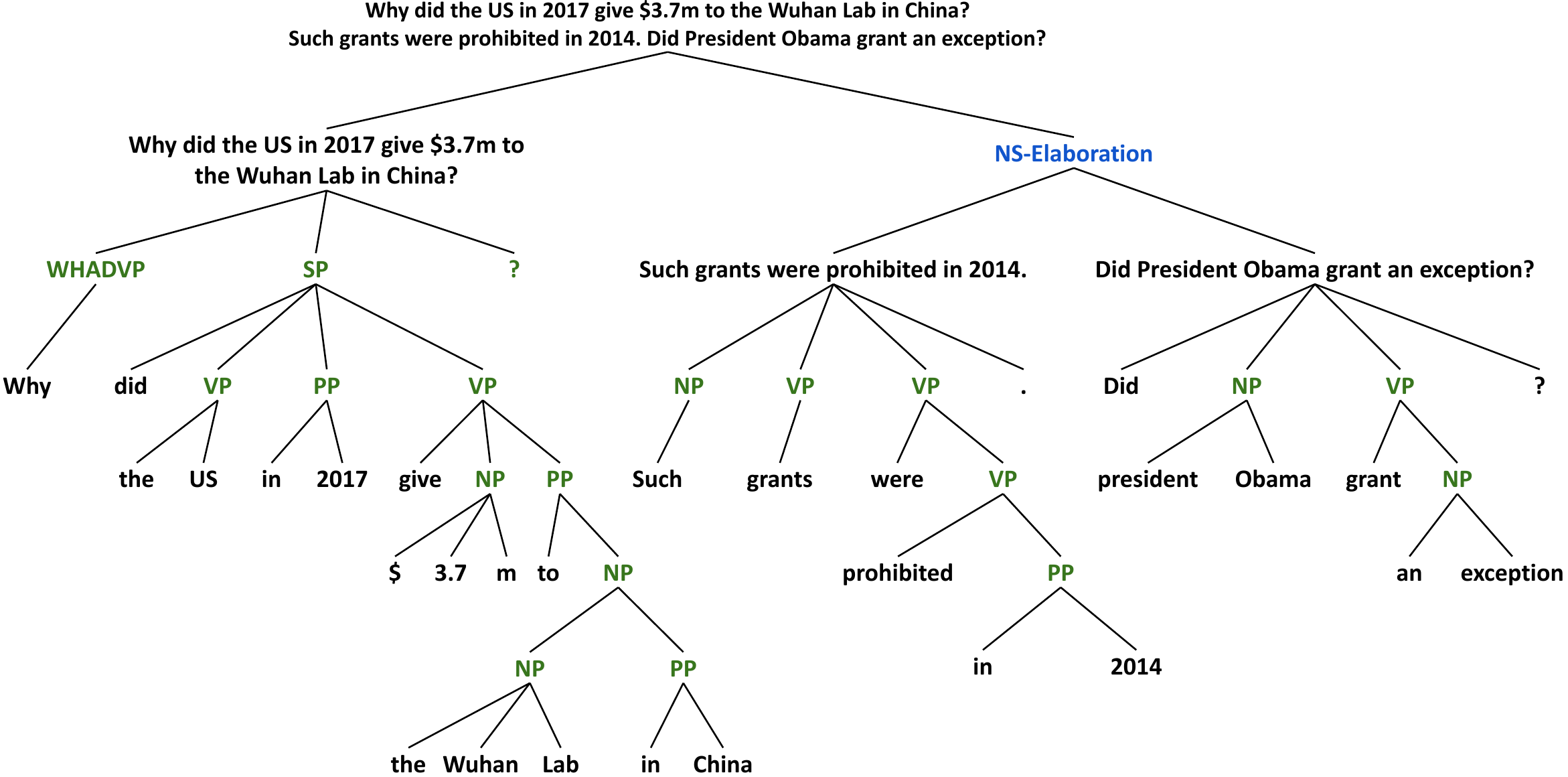}}
    \caption{The hierarchical linguistic tree for the news piece ``Why did the US in 2017 give \$3.7m to the Wuhan Lab in China? Such grants were prohibited in 2014. Did President Obama grant an exception?'' verified as a false statement by PolitiFact.\protect\footnotemark Blue nodes: RRs. Green nodes: POSs.}
    \label{fig:hierarchical_linguistic_tree}
\end{figure*}

These studies have advanced linguistic-style-aware fake news prediction. However, translating a news article's linguistic style into the appearances of words, POSs, and RRs overlooks the linguistic structure that reveals how the article's words, POSs, and RRs are assembled. Specifically, we can form a hierarchical linguistic tree for each news article; see Section \ref{subsec:hlt_construction} for the details and Figure~\ref{fig:hierarchical_linguistic_tree} for an illustrated tree for the news piece ``\texttt{Vitamin D determines severity in COVID-19, so government advice needs to change, experts urge: Researchers point to changes in government advice in Wales, England, and Scotland.}'' This tree explicitly presents {\it the order of words} used in the article, {\it syntactic structure} revealing how these words are recursively structured as the elementary discourse units (EDUs, which are meaningful phrases, sentences, or paragraphs) through POSs, and discourse structure exhibiting how these EDUs are recursively structured as the entire article through RRs. Previous approaches paid full attention to the tree's {\it node} information by looking if this news piece used a specific word (e.g., ``COVID-19''), POS (e.g., ``NNP''), or RR (e.g., ``NS-elaboration'') in the corpus or how many times it appears without considering the {\it relational (edge)} information among the nodes. Although Zhou et al.~\cite{zhou2020recovery} and P{\'e}rez-Rosas et al.~\cite{perez2018automatic} computed the frequencies of production rules (at the syntactic level only), each rule can merely show the structure within a fundamental component of the tree (i.e., parent-children, such as VP $\rightarrow$ VBZ NP). Each fundamental component is investigated independently by overlooking how components are connected to form the tree; the tree's structure is hence preserved {\it locally} rather than {\it globally}. In addition, the representation of news articles obtained by the frequencies of these local structures is often high-dimensional and sparse, which can be adverse to the prediction task.

\footnotetext{Source: \url{https://www.politifact.com/factchecks/2020/may/01/rudy-giuliani/rudy-giuliani-wrong-about-us-policy-grant-amount-w/}}

\vspace{6pt}
\noindent\textbf{Present work.} 
To address the above problems, we propose the hierarchical recursive neural network (HERO) for fake news prediction. The architecture of the proposed neural network adaptively preserves the global structure of the hierarchical linguistic tree of various news articles. To our best knowledge, this is the first work that develops hierarchical linguistic trees. Leveraging the developed trees, the proposed neural network can learn the linguistic-style-aware representations of news articles by explicitly capturing the writers' usage of words and the linguistically meaningful ways in which these words are structured as phrases, sentences, paragraphs, and, ultimately, the documents. We conduct extensive experiments on real-world datasets with well-established and state-of-the-art approaches, which demonstrate the effectiveness of the proposed neural network in predicting fake news. Additionally, we examine the differential linguistic style of fake news and the truth and identify statistically significant and consistent patterns of fake news across datasets.

\vspace{6pt}
The rest of this paper is organized as follows. We review related work in Section \ref{sec:related_work}. We introduce the proposed method in Section \ref{sec:method} and detail the experiments designed and conducted to evaluate the proposed method in Section \ref{sec:experiments}. We conclude in Section \ref{sec:conclusion}.

\begin{figure*}
    \centering
    \includegraphics[width=\textwidth]{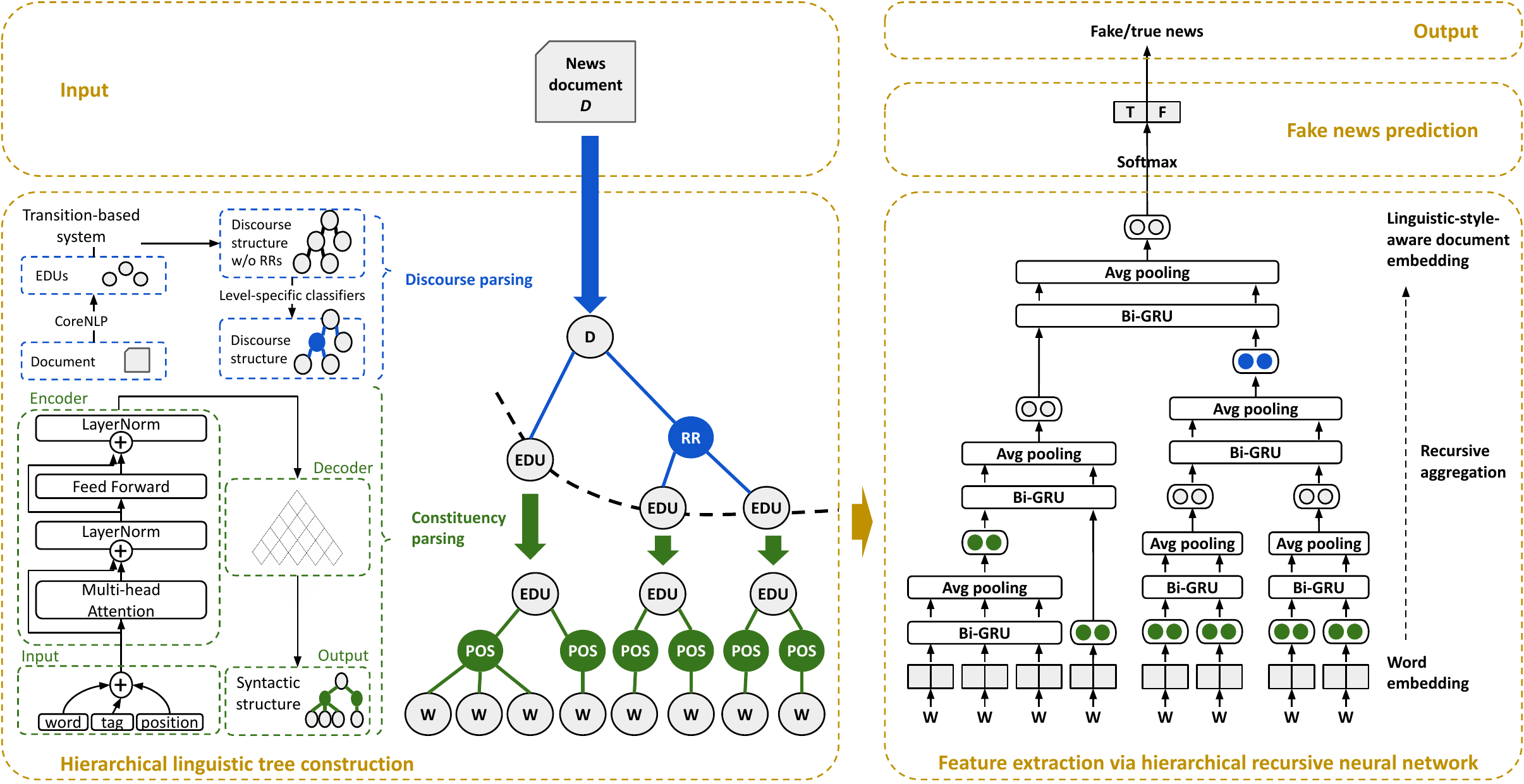}
    \caption{Framework overview, which contains a top-bottom building process of hierarchical linguistic trees, a bottom-top feature extraction process using the proposed hierarchical recursive neural network, and a classifier to predict fake news. The neural network's architecture adaptively preserves various news documents' global and hierarchical linguistic tree structures. The Bi-GRU aggregator catches text's local sequentiality that is linguistically valuable and often short (explained in Section \ref{subsec:hero}), which is more effective than self-attention here (see Section \ref{subsubsec:best_hero} for details).}
    \label{fig:framework_overview}
\end{figure*}

\section{Related Work}
\label{sec:related_work}

Fake news prediction methods can be categorized as content-based or propagation-based depending on whether the method focuses on investigating news content or its propagation on social media. 

Propagation-based methods can utilize rich auxiliary social media information, including news spreaders' 
intent~\cite{zhou2022fake} or
profiles~\cite{cheng2021causal}, relationships between news spreaders and their posts~\cite{min2022divide}, social feedback~\cite{shu2019defend,qian2018neural,zhang2019reply}, social networks~\cite{tommasel2022following}, and propagation paths~\cite{naumzik2022detecting,liu2018early}. Nevertheless, they can only be deployed after news articles published on news outlets have been disseminated on social media. In comparison, content-based methods have the primary advantage of predicting fake news early when news articles have been published online but have not been spread~\cite{zhou2020fake}. Additionally, an effective content-based method can be easily extended further by incorporating social context information. With this consideration, we focus on analyzing news content to predict fake news and review related work on content-based fake news prediction approaches.

As news articles are mainly text, content-based methods start by manually extracting linguistic features and predicting fake news using common classifiers such as SVM~\cite{perez2018automatic}. Such linguistic features have been related to lexicons (e.g., bag-of-words)~\cite{zhou2020fake}, POSs~\cite{potthast2018stylometric,zhou2020fake}, context-free grammars (production rules)~\cite{perez2018automatic,zhou2020fake}, RRs~\cite{rubin2015truth,zhou2020fake}, readability~\cite{potthast2018stylometric,zhu2022memory}, and $n$-grams that preserve the sequences of words or POSs~\cite{przybyla2020capturing}. Though news features can be easily interpreted within this machine learning framework, features cannot be automatically extracted, which can significantly impact the prediction performance; hence, the performance heavily relies on experts' involvement and experience. More importantly, as detailed in Section \ref{sec:introduction}, it is difficult to capture the global structure of news text (language) at any of the syntactic and discourse levels with these hand-crafted features. Compared to these methods, the proposed neural network can learn the features of news articles, which capture the global and hierarchical structures that news linguistic styles carry.

Recently, neural networks (e.g., Bi-LSTM~\cite{przybyla2020capturing,karimi2019learning} and Text-CNN \cite{wang2018eann}) have been frequently employed to identify fake news. These models can learn the features of news text (sometimes, combined with other modalities in news content, such as images~\cite{wang2018eann,cui2019same}).
These neural networks have focused on the sequentiality or locality of news text but not on its linguistic structure. In comparison, the proposed neural network explicitly catches this structure; it also captures text's sequentiality and locality, which will be detailed in Section \ref{subsec:hero}. We point out that the proposed neural network provides a fundamental approach to news text representation learning and thus can be easily extended for multimodal fake news prediction. 

\section{Methodology}
\label{sec:method}

We specify the proposed model in this section, which can be divided into three steps. For each news document, we first construct its
hierarchical linguistic tree (see Section~\ref{subsec:hlt_construction}), then extract its features via the proposed hierarchical recursive neural network that preserves the hierarchical linguistic tree information (see Section \ref{subsec:hero}),
and finally predict it as fake news or the truth (see Section \ref{subsec:fake_news_prediction}).
Figure 2 presents the framework overview.

\subsection{Hierarchical Linguistic Tree Construction}
\label{subsec:hlt_construction}

Given a news document $D$, we first generate its hierarchical linguistic tree. The tree can explicitly present the order of words used in the document and how these words shape EDUs (meaningful phrases, sentences, or paragraphs) and further shape the entire document. An example is shown in Figure~\ref{fig:hierarchical_linguistic_tree}.
Specifically, our first attempt is to obtain $D$'s discourse (rhetorical) structure, which identifies $D$'s EDUs and reveals how these EDUs recursively form the document $D$. To this end, we first utilize Standford CoreNLP~\cite{manning2014stanford} to segment $D$ into EDUs. Then, we apply a modified transition-based system~\cite{wang2017two} to identify span ({S}) and nuclearity ({N}), based on which $D$'s rhetorical structure can be obtained without recognizing specific RRs (e.g., \textit{elaboration} in Figure~\ref{fig:hierarchical_linguistic_tree}). This semi-naked tree structure allows us to divide each RR node into within-sentence, across-sentence, and across-paragraph levels in terms of its left and right subtrees, extract structural features for each RR node, and ultimately adopt level-specific SVM classifiers~\cite{wang2017two} to predict the node attribute (i.e., the specific RR). This multi-stage approach outperforms the well-established one~\cite{ji2014representation} in our experiments, where \cite{ji2014representation} works as an integrated system. Finally, we employ a state-of-the-art discriminative constituency parser for each identified EDU of the document $D$ to obtain its syntactic structure~\cite{kitaev2018constituency}. The parser consists of a self-attentive encoder~\cite{kitaev2018constituency} and a chart decoder~\cite{gaddy2018s} (see Figure~\ref{fig:framework_overview} for the detailed architecture). The syntactic structure reveals how the EDU's words recursively form the entire EDU.

\subsection{\small Feature Extraction via Hierarchical Recursive Neural Network}
\label{subsec:hero}

We propose the hierarchical recursive neural network to extract features of news documents, whose architecture adaptively maintains the global structure of the hierarchical linguistic trees of news documents. 

Given a news document $D$, its feature extraction using the hierarchical recursive neural network is bottom-top. We first encode $D$'s words, which are the leaf nodes of $D$'s hierarchical linguistic tree.
Then, we aggregate the obtained embeddings of the words that attach to the same parent node, forming the embedding of their parent node. The hierarchical recursive neural network will repeat such aggregations from the lower (syntactic) level to the upper (discourse) level until the document $D$ as the tree's root node is embedded.

Hence, the question arises of how the aggregator performs on a recurring fundamental component (i.e., a depth-one parent-children structure). Note that for each parent node in a hierarchical linguistic tree, its children contain \textit{local} and \textit{sequential} information of the corresponding news document. The information is local because it reveals partial information about the overall news content that is linguistically valuable. It is sequential as we keep the order of words of the news document. Naturally, recurrent neural networks can be adopted as the aggregator to catch the sequentiality of children sharing the same parent. The information locality further relieves the pressure on recurrent neural networks to keep the dependency of long entities since the number of children for a parent node is no more than the EDU length and essentially less than the document length. As seen from Figure~\ref{fig:hierarchical_linguistic_tree}, the maximum length that recurrent neural networks require to process is four, whereas the document has 29 tokens.

With the above considerations, we develop Bi-GRU (bidirectional gated recurrent unit), one of the well-established recurrent neural networks~\cite{cho2014learning}, to aggregate the embeddings of all the child nodes to represent their parent node. We also empirically compare Bi-GRU with multi-head self-attention that has remarkably performed in many tasks; Bi-GRU is more effective for the proposed model. Formally, the embedding of a parent node is computed as
\begin{equation}\label{eq:aggregator}
    \textbf{x}_p = \frac{\sum_{c\in\mathcal{C}_p} [\overrightarrow{GRU}\{ \textbf{x}_c\} \oplus \overleftarrow{GRU}\{ \textbf{x}_c\}]}{|\mathcal{C}_p|}, 
\end{equation}
where $p$ denotes the parent node and $\mathcal{C}_p$ is the set of child nodes of $p$. Vectors $\mathbf{x}_c \in \mathbb{R}^{d}$ and $\mathbf{x}_p \in \mathbb{R}^{d}$ refer to the features of the child and parent node, respectively. The operator $\oplus$ denotes concatenation. The GRU is formulated as follows:
\begin{equation}\label{eq:bi_gru}
\begin{array}{l}
\mathbf{r}_{i} = \sigma(\mathbf{W}_r \mathbf{x}_i + \mathbf{U}_r\mathbf{h}_{i-1}), \\ 
\mathbf{z}_i = \sigma(\mathbf{W}_z \mathbf{x}_i + \mathbf{U}_z\mathbf{h}_{i-1}), \\ 
\hat{\mathbf{h}}_i = \tanh(\mathbf{W}_h \mathbf{x}_i + \mathbf{U}_h(\mathbf{h}_{i-1}\odot \mathbf{r}_i)), \\ 
\mathbf{h}_i = (1-\mathbf{z}_i) \odot \mathbf{h}_{i-1} + \mathbf{z}_i \odot \hat{\mathbf{h}}_i,
\end{array}
\end{equation}
where $\mathbf{h}_i\in\mathbb{R}^{d/2}$ is the output hidden state of the $i$-th child, with $\mathbf{h}_0 = \mathbf{0}$. The symbol $\odot$ denotes Hadamard product. Matrices $\mathbf{W}_* \in\mathbb{R}^{(d/2)\times d}$ and $\mathbf{U}_*\in\mathbb{R}^{(d/2)\times (d/2)}$ ($*\in \{r,z,h \}$) are learnable parameters. $\mathbf{r}_i$ and $\mathbf{z}_i$ are the reset gate and update gate, respectively. $\sigma$ and $\tanh$ are the sigmoid and hyperbolic tangent activation functions, respectively.
In a nutshell, the above architecture first employs the Bi-GRU to capture ``deep'' sequential feature interactions of all the child features and then uses a mean pooling layer over all the hidden states to obtain the parent node's features.  

After determining the aggregation within each recurring fundamental component, we introduce three specific hierarchical recursive neural networks (HEROs):

\begin{itemize}
    \item \textit{Unified HERO}: The first hierarchical recursive neural network is the one with unified aggregators. In other words, all the Bi-GRUs in the neural network share the same set of $\mathbf{W}_*$ and $\mathbf{U}_*$ ($*\in\{r,z,h \}$, see Equation (\ref{eq:bi_gru})). 
    
    \item \textit{Level-specific HERO}: It is the hierarchical recursive neural network with level-specific aggregators. As detailed, the hierarchical linguistic tree presents both syntactic and rhetorical structures of news content, and the hierarchical recursive neural network preserves such structures. Hence, all the Bi-GRUs in a hierarchical recursive neural network can be grouped by the level (syntax or discourse) they belong to the corresponding tree. We define $L(v)$ as the function that maps a certain vertex in the hierarchical linguistic tree to its linguistic level (i.e., $L(v)\in \{\text{syntax}, \text{discourse} \}$). Then, Equation (\ref{eq:aggregator}) can be reformulated as $\textbf{x}_p = \frac{1}{|\mathcal{C}_p|}\sum_{c\in\mathcal{C}_p} [\overrightarrow{GRU}_{L(c)}$ $\{ \textbf{x}_c\} \oplus \overleftarrow{GRU}_{L(c)}\{ \textbf{x}_c\}]$.
    
    \item \textit{Attribute-specific HERO}: It stands for the hierarchical recursive neural network with attribute-specific aggregators. In other words, we categorize the hierarchical recursive neural network's recurring fundamental components according to the attributes of their parent nodes in the corresponding hierarchical linguistic tree, which can be various POSs and RRs. We deploy the same Bi-GRU for the components within each category and the different Bi-GRU for the components falling into different categories. Mathematically, we define $A(v)$ as the function that maps a certain vertex in the hierarchical linguistic tree to its attributes. Assume there are $m$ different POSs  and $n$ RRs, we have  $A(v)\in \{POS_i, RR_j: i=1,2,\cdots,m, j=1,2,\cdots,n\}$. The root vertex would be assigned with a RR in discourse parsing. For EDU vertices, they would not be assigned with any RRs in discourse parsing but with some POSs in constituency parsing. In this way, Equation (\ref{eq:aggregator}) is rewritten as $\textbf{x}_p = \frac{1}{|\mathcal{C}_p|}\sum_{c\in\mathcal{C}_p} [\overrightarrow{GRU}_{A(p)}\{ \textbf{x}_c\} \oplus \overleftarrow{GRU}_{A(p)}\{ \textbf{x}_c\}]$. 
\end{itemize}

\subsection{Fake News Prediction}
\label{subsec:fake_news_prediction}

We add a softmax classifier on the top of the proposed hierarchical recursive neural network to predict the document $D$ as fake news or the truth. Let $\mathbf{h}_D$ denote $D$'s features extracted via the proposed hierarchical recursive neural network. The softmax function maps $\mathbf{h}_D$ to the probability of $D$ being a fake news document by
\begin{math}
    p_D = \text{Softmax}(\mathbf{W} \mathbf{h}_D+ \mathbf{b}),
\end{math}
where $\mathbf{W}$ and $\mathbf{b}$ are learnable parameters.

\vspace{6pt}
To learn the parameters $\Theta = \{ \mathbf{W}_*, \mathbf{U}_*, \mathbf{W}, \mathbf{b} \}$ within the neural network and classifier, we employ cross-entropy to calculate the classification loss in the model training process. Assume we have $q$ verified news documents $\mathcal{D}=\{D_i\}_{i=1}^q$ with the ground-truth labels $\mathcal{Y}=\{y_i: y_i\in\{0,1\}\}_{i=1}^q$ ($y_i=0$ for true news, and $y_i=1$ for fake news), the loss is computed by $L = -\frac{1}{q} \sum_{i=1}^q [y_{D_i} \log p_{D_i} + (1-y_{D_i}) \log (1-p_{D_i})]$.
Based on it, the parameter set $\Theta$ is estimated by
\begin{math}
    \hat{\Theta} = \arg \min_{\Theta} L.
\end{math}

\section{Empirical Evaluation}
\label{sec:experiments}

We aim to evaluate the proposed method by answering the following three questions.
\begin{enumerate}
    \item How effective is the proposed model in fake news prediction compared to the-state-of-art approaches?
    \item Is the hierarchical linguistic structure of news documents essential in representing their linguistic styles? 
    \item What characterizes the linguistic style of fake news as distinguishable from the truth?
\end{enumerate}
To that end, we first detail our experimental setup in Section~\ref{subsec:experiment_setup} and then compare the proposed unified, level-specific, and attribute-specific HEROs in predicting fake news (see Section \ref{subsubsec:best_hero}). Subsequently, we compare the proposed model with the baselines to verify its effectiveness in predicting fake news (to answer RQ1, see Section \ref{subsubsec:hero_vs_baselines}) and conduct the ablation study to assess the importance of our developed hierarchical linguistic trees (to answer RQ2, see Section \ref{subsubsec:ablation_study}). Finally, we characterize the linguistic style of fake news as distinguishable from the truth by doing quantitative and comparative analyses (to answer RQ3, see Section \ref{subsubsec:fake_news_patterns}).

\subsection{Experimental Setup}
\label{subsec:experiment_setup}

We first introduce the datasets used for evaluation (see Section \ref{subsubsec:datasets}), followed by the baselines for comparison (see Section \ref{subsubsec:baselines}). Finally, we detail our implementation details in Section \ref{subsubsec:implementation_details}.

\subsubsection{Datasets}
\label{subsubsec:datasets}

We conduct experiments on two benchmark datasets in fake news prediction: Recovery~\cite{zhou2020recovery} and MM-COVID~\cite{li2020mm}. Both datasets contain labeled news documents. Differently, news documents collected in Recovery are articles (long text, often including multiple paragraphs) but in MM-COVID are statements (short text, often formed by one or two sentences). We present the detailed statistics of two datasets in Table~\ref{tab:data_statistics}.

\begin{table}[t]
\centering
\caption{Data statistics.}
\label{tab:data_statistics}
\begin{tabular}{lrr}
\toprule[1pt]
 & \textbf{Recovery} & \textbf{MM-COVID} \\ \midrule
\textbf{\# news documents} & 2,029 & 3,536 \\
\textbf{\quad - true news} & 1,364 & 1,444 \\
\textbf{\quad - fake news} & 665 & 2,092 \\
\textbf{Avg. \# words per EDU} & 24 & 17 \\
\textbf{Avg. \# EDUs per document} & 38 & 2 \\
\textbf{Avg. \# words per document} & 841 & 16 \\
\bottomrule[1pt]
\end{tabular}
\end{table}

\subsubsection{Baselines}
\label{subsubsec:baselines}

We involve the following well-received and state-of-the-art methods as baselines in our experiments.
\begin{itemize}
    \item \textit{HCLF}~\cite{zhou2020fake}: HCLF stands for hand-crafted linguistic feature. Each news document's HCLFs include the frequencies of words (i.e., bag-of-word features), POSs, RRs, and production rules. The extracted features are used to predict fake news by employing well-established classifiers. Here we examine a comprehensive list of classifiers--logistic regression, SVM, $k$-nearest neighbors, decision trees, naive Bayes, random forest, and AdaBoost--and select the one performing best.

    \item \textit{EANN}~\cite{wang2018eann}: The event adversarial neural network contains three components: feature extraction by Text-CNN (for text) and VGG-19 (for images), event discrimination to learn event-invariant features of news content, and fake news prediction. We exclude the visual features for a fair comparison.
    
    \item \textit{HAN}~\cite{yang2016hierarchical}: HAN exploits attention-GRU for news classification. It captures the hierarchical sequence of documents; i.e., each document is a sequence of its sentences, and each of its sentences is a sequence of words. 
    
    \item \textit{DRNN}~\cite{ji2017neural}: DRNN is a discourse-structure-aware neural network, which focuses on the tree with rhetorical relationships as edge attributes and leverages an attention mechanism for news classification. In other words, DRNN differs from HERO in the aggregation rule. Compared to DRNN's tree, the hierarchical linguistic tree integrates syntax-level structures and has RRs as nodes and non-attributed edges at the discourse level. DRNN is developed to categorize news documents with more than one elementary discourse unit; otherwise, it is reduced to Bi-LSTM.
    
    \item \textit{Text-GCN}~\cite{yao2019graph}: The approach develops the graph convolutional neural network for news classification. The graph investigates the co-occurrence relationship among news documents and the words within the documents. 
    
    \item \textit{Transformer}~\cite{vaswani2017attention}: It is a deep neural network model with a self-attention-based encoder-decoder architecture, which has excellently performed in diverse natural language processing tasks. Here, we consider Transformer's encoder--applicable for classification tasks--as a baseline to predict fake news, a non-pretrained version rather than pretrained models (e.g., BERT) for a fair comparison as the pretrained Transformers have learned large-scale external resources.
\end{itemize}

\subsubsection{Implementation Details}
\label{subsubsec:implementation_details}

We randomly divide each dataset into 0.7:0.1:0.2 proportions for model training, validation, and testing. Macro-F1, micro-F1, and AUC are used to evaluate the performance of methods in news classification. The discourse parser is pretrained using RST-DT~\cite{wang2017two}, and the constituency parser is pretrained using the Penn Treebank~\cite{kitaev2018constituency}.
For the neural-network-based models, we uniformly utilize the pretrained GloVe~\cite{pennington2014glove} to obtain semantic-aware embeddings of words, with 100 as the embedding dimension. The hidden dimension within neural networks is set as 100, correspondingly. 
We deploy Adam optimizer to learn parameters, with 50 as the maximum number of epochs. We perform a grid search over the learning rate $\in \{0.1, 0.01, 0.001, 0.0001\}$ with validation data. In the end, 0.0001 performs best for our models and most of the baselines other than Transformer (0.001) and Text-GCN (0.01). All the experiments of the neural networks are implemented with PyTorch and are conducted on an NVIDIA Quadro RTX 6000 GPU ($24$ GB memory), Intel(R) Xeon(R) Gold 6248R CPU ($3.00$ GHz), and with $64$ GB of RAM. For HCLFs, classifiers are used with the default hyperparameters presented in the scikit-learn library. $Z$-score normalization is applied for the feature matrix to enhance the classification performance.

\subsection{Determining the Best HERO} 
\label{subsubsec:best_hero}

We compare the performance of the proposed neural networks with unified, level-specific, and attribute-specific Bi-GRUs in predicting fake news. Table~\ref{tab:performance_three_heros} presents the results. The results indicate that 
with Recovery data, the performance ranking is attribute-specific HERO $>$ level-specific HERO $>$ unified HERO. Specifically, attribute-specific HERO correctly predicts news as fake or true with 0.85 macro-F1 and 0.87 micro-F1 and AUC, outperforming unified HERO by $\sim$4\% and level-specific HERO by $\sim$3\%. 
With MM-COVID data, the performance ranking is attribute-specific HERO $\approx$ unified HERO $>$ level-specific HERO. Attribute-specific and unified HEROs achieve $\sim$0.89--0.90\% in macro-F1, micro-F1, and AUC, outperforming level-specific HERO by $\sim$1\%.
In conclusion, {\it attribute-specific HERO} performs best in classifying long articles and short statements as fake news or the truth. This result demonstrates the importance of the node attributes (POSs or RRs) in developed hierarchical linguistic trees.

Additionally, we compare Bi-GRU and self-attention (\#heads=10) as aggregators in the proposed hierarchical recursive neural network for fake news prediction. The results indicate that Bi-GRU performs better than self-attention by at least 1\% in AUC on both datasets.

\begin{table}[t]
    \centering
    \caption{Performance of unified, level-specific, and attribute-specific HEROs in fake news prediction. Attribute-specific HERO performs best, demonstrating that the node attributes (POSs or RRs) in hierarchical linguistic trees are essential. MAF1: Macro-F1. MIF1: Micro-F1}
    \label{tab:performance_three_heros}
    \begin{adjustbox}{width=\columnwidth}
        \begin{tabular}{rcccccc}
        \toprule[1pt]
         & \multicolumn{3}{c}{\textbf{Recovery}} & \multicolumn{3}{c}{\textbf{MM-COVID}} \\ \cmidrule[0.5pt]{2-7}
        \textbf{HERO} & \textbf{MAF1} & \textbf{MIF1} & \textbf{AUC} & \textbf{MAF1} & \textbf{MIF1} & \textbf{AUC} \\ \midrule[0.5pt]
        \textbf{Unified} & 0.801 & 0.822 & 0.827 & 0.889 & 0.891 & \textbf{0.899} \\
        \textbf{Level-specific} & 0.817 & 0.838 & 0.841 & 0.878 & 0.878 & 0.892 \\
        \textbf{Attribute-specfic} & \textbf{0.850} & \textbf{0.869} & \textbf{0.866} & \textbf{0.894} & \textbf{0.896} & 0.896 \\
        \bottomrule[1pt]
        \end{tabular}
    \end{adjustbox}
\end{table}

\begin{table}[t]
    \centering 
    \caption{Performance of the proposed model, HERO, and baselines in fake news prediction. HERO outperforms the baselines by 2--17\% in AUC on Recovery and by 3--30\% in MM-COVID. MAF1: Macro-F1. MIF1: Micro-F1}
    \label{tab:hero_vs_baselines}
    % \begin{adjustbox}{width=\columnwidth}
    \begin{tabular}{rcccccc}
    \toprule[1pt]
    \multirow{2}{*}{} & \multicolumn{3}{c}{\textbf{Recovery}} & \multicolumn{3}{c}{\textbf{MM-COVID}} \\ \cmidrule[0.5pt]{2-7}
     & \textbf{MAF1} & \textbf{MIF1} & \textbf{AUC} & \textbf{MAF1} & \textbf{MIF1} & \textbf{AUC} \\ \midrule[0.5pt]
    \textbf{HCLF} & 0.752 & 0.801 & 0.746 & 0.566 & 0.624 & 0.577 \\
    \textbf{Transformer} & 0.774 & 0.793 & 0.810 & 0.804 & 0.809 & 0.806 \\
    \textbf{Text-GCN} & 0.841 & 0.869 & 0.835 & 0.826 & 0.836 & 0.817 \\
    \textbf{EANN} & 0.811 & 0.864 & 0.795 & 0.825 & \textbf{0.926} & 0.833 \\
    \textbf{HAN} & 0.847 & 0.869 & 0.844 & 0.840 & 0.856 & 0.846 \\
    \textbf{DRNN} & 0.711 & 0.778 & 0.698 & 0.845 & 0.846 & 0.848 \\
    \textbf{HERO} & \textbf{0.850} & \textbf{0.869} & \textbf{0.866} & \textbf{0.894} & 0.896 & \textbf{0.896} \\
    \bottomrule[1pt]
    \end{tabular}
% \end{adjustbox}
\end{table}

\subsection{Comparing HERO with Baselines}
\label{subsubsec:hero_vs_baselines}

We compare the proposed model with the baselines in predicting fake news. The results presented in Table~\ref{tab:hero_vs_baselines} reveal that the proposed model can generally outperform the baselines. Specifically, the proposed model has an AUC score approaching 0.87, outperforming 
HAN by more than 2\%, 
Text-GCN by more than 3\%, 
Transformer by more than 5\%, 
EANN by more than 7\%, 
HCLF by 12\%, and 
DRNN by 17\%. 
With MM-COVID data, the proposed model has an AUC score approaching 0.90, outperforming 
EANN by more than 6\%, 
DRNN and HAN by $\sim$5\%, 
Text-GCN and Transformer by $\sim$8-9\%, 
and HCLF by more than 30\%. 
From the table, we also observe that the proposed model outperforms EANN by 6--7\% in macro-F1 and AUC but underperforms it by $\sim$3\% in micro-F1 on MM-COVID. This result suggests that EANN tends to predict given news statements as the major class.

\begin{figure}[t]
    \centering
    \subfloat[Recovery]{
        \includegraphics[width=0.47\columnwidth]{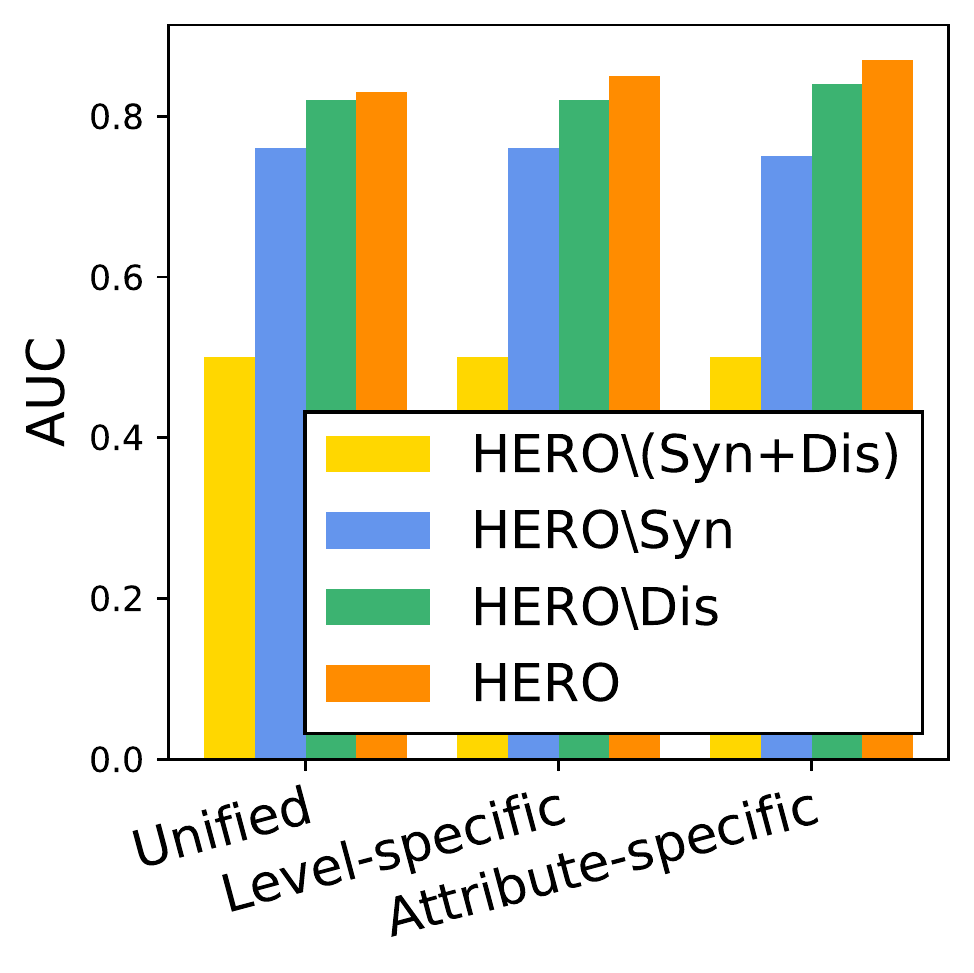}
    }
    \subfloat[MM-COVID]{
        \includegraphics[width=0.47\columnwidth]{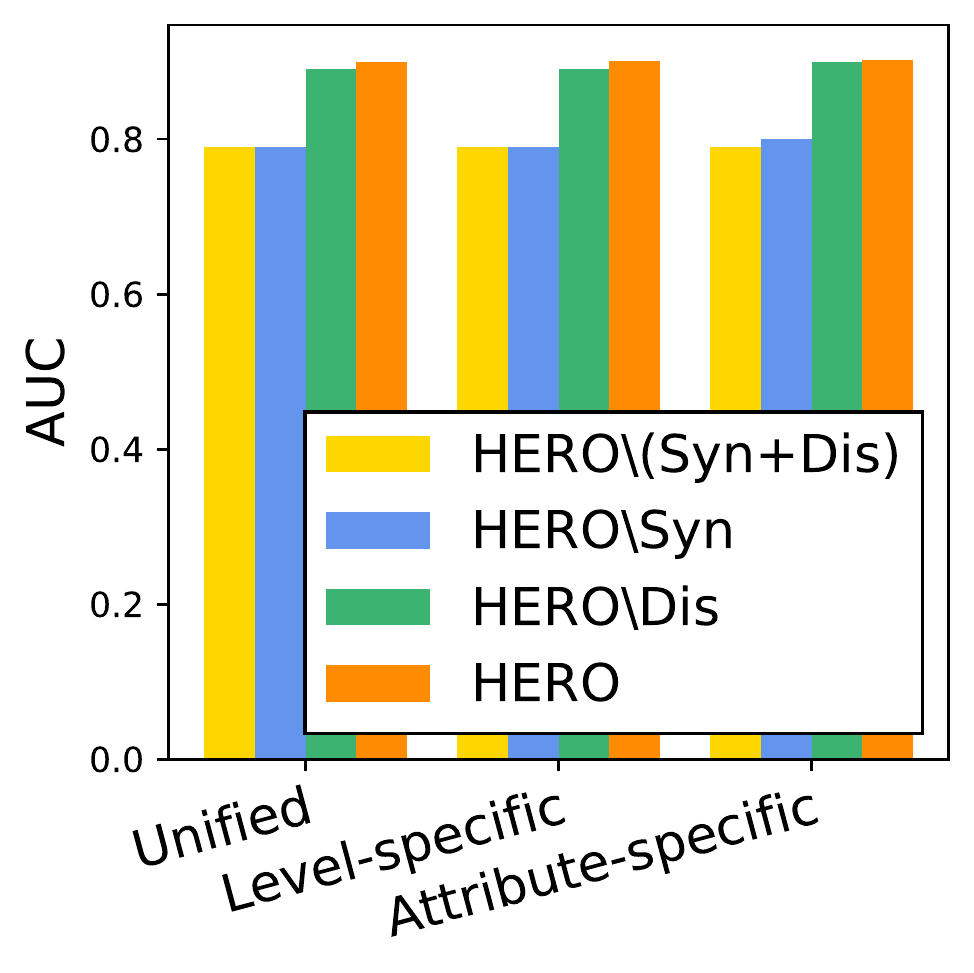}
    }
    \caption{Ablation study. (a) The proposed HERO outperforms HERO$\setminus$Dis by 1\% in AUC for unified HERO and by 3\% for level- and attribute-specific HEROs. It outperforms HERO$\setminus$Syn by 7--9\% and HERO$\setminus$Syn by 30\%+ in AUC. (b) HERO performs similarly to HERO$\setminus$Dis as MM-COVID contains short statements having minimal discourse structures (i.e., syntax-level structures dominate). It outperforms HERO$\setminus$Syn and HERO$\setminus$(Syn+Dis) by 10\%+ in AUC. Thus, syntax- and discourse-level structures are both essential.}
    \label{fig:ablation_study}
\end{figure}

\begin{figure*}
    \centering
    \subfloat[Children of parent nodes.]{\label{subfig:children_of_parent_nodes}
    \begin{minipage}[b]{0.24\textwidth}\centering
        \includegraphics[width=0.54\columnwidth]{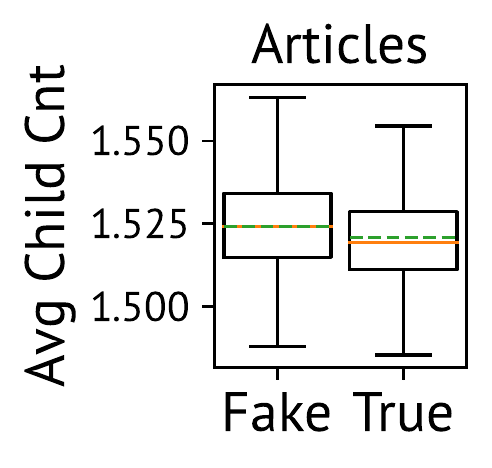}
        \includegraphics[width=0.43\columnwidth]{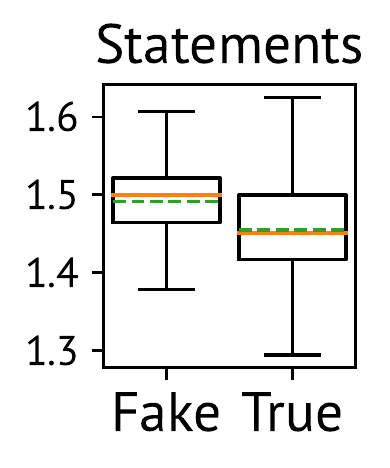}    
    \end{minipage}
    } 
    \subfloat[Note attributes.]{\label{subfig:node_attributes}
    \begin{minipage}[b]{0.74\textwidth}\centering
        \includegraphics[width=0.175\columnwidth]{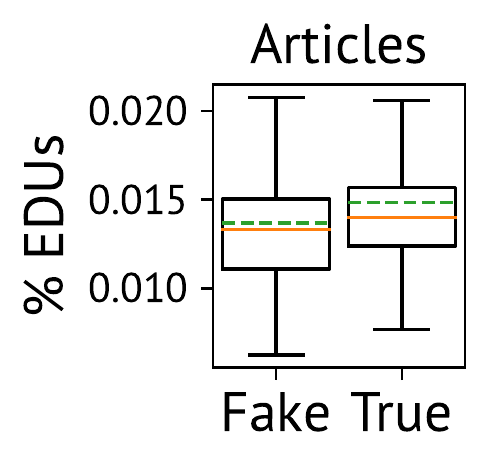}
        \includegraphics[width=0.146\columnwidth]{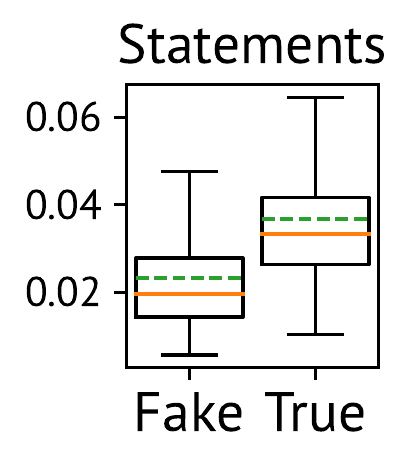}  
        \includegraphics[width=0.166\columnwidth]{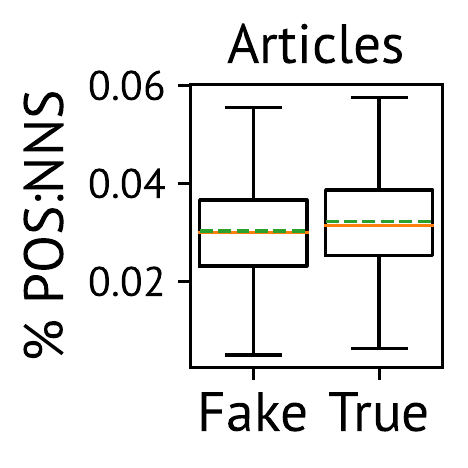}
        \includegraphics[width=0.145\columnwidth]{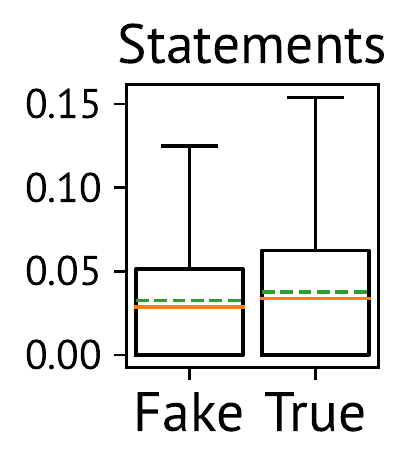}   
        \includegraphics[width=0.166\columnwidth]{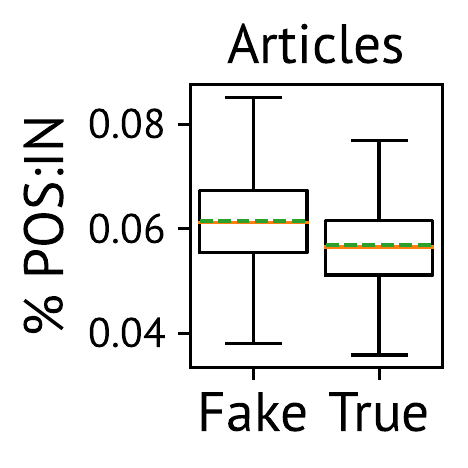}
        \includegraphics[width=0.145\columnwidth]{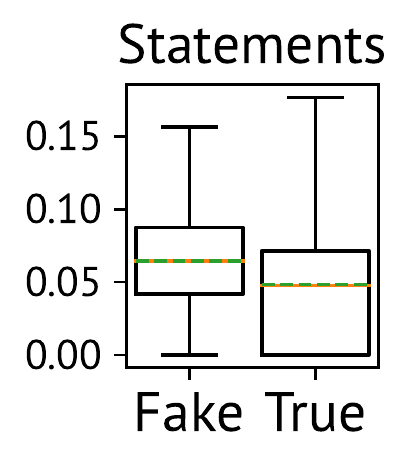}   
    \end{minipage}
    }    
    
    \begin{minipage}{\textwidth}
        \includegraphics[width=0.13\columnwidth]{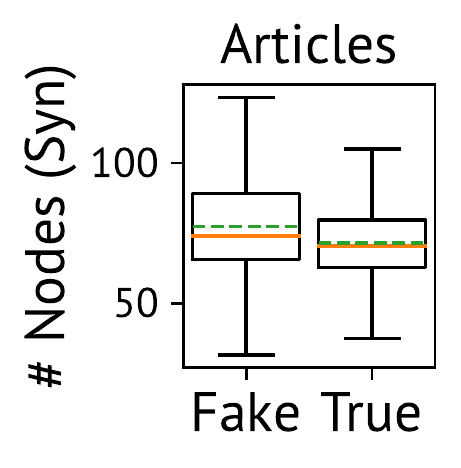}
        \includegraphics[width=0.1125\columnwidth]{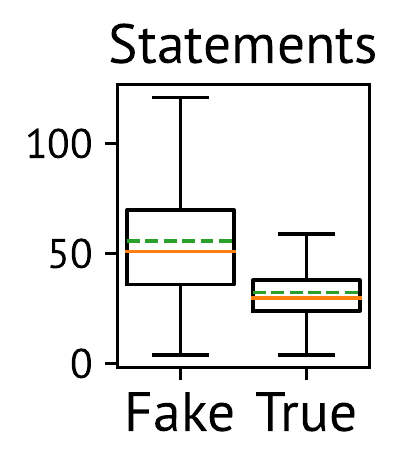}
        \includegraphics[width=0.125\columnwidth]{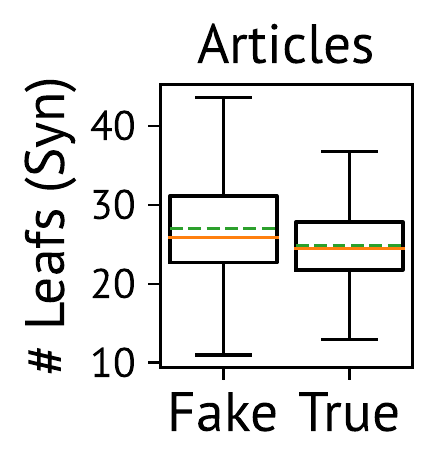}
        \includegraphics[width=0.1075\columnwidth]{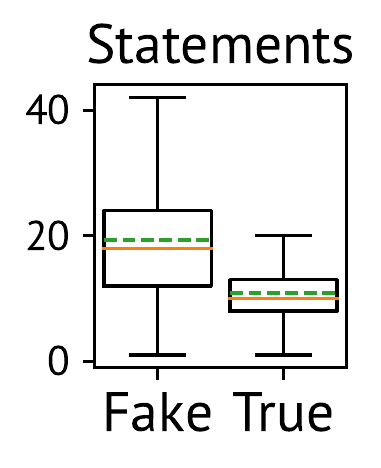}
        \includegraphics[width=0.125\columnwidth]{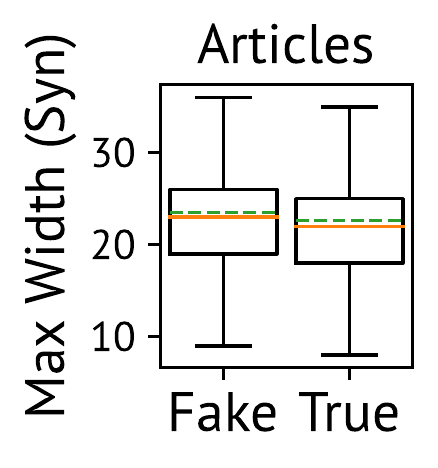}
        \includegraphics[width=0.1075\columnwidth]{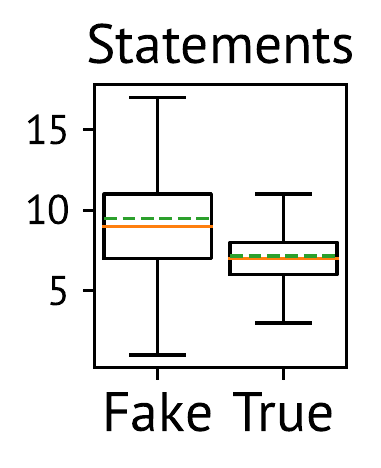}
        \includegraphics[width=0.125\columnwidth]{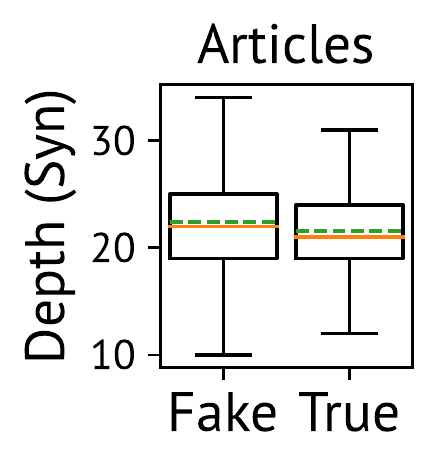}
        \includegraphics[width=0.105\columnwidth]{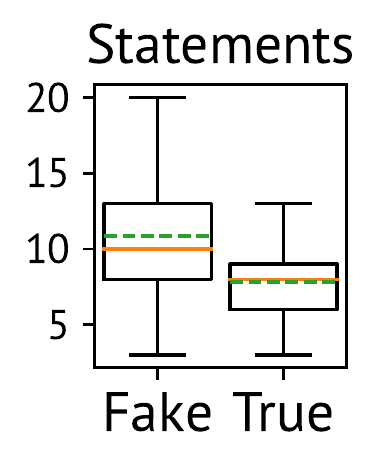}
    \end{minipage}

    \subfloat[Size, width, and depth of trees.]{  \label{subfig:size_of_trees}
    \begin{minipage}{\textwidth}\centering
        \includegraphics[width=0.135\textwidth]{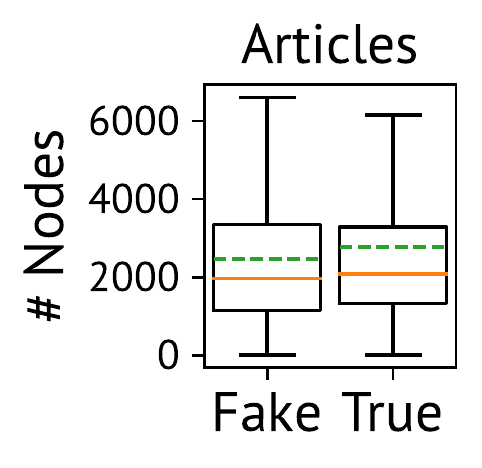} 
        \includegraphics[width=0.1275\textwidth]{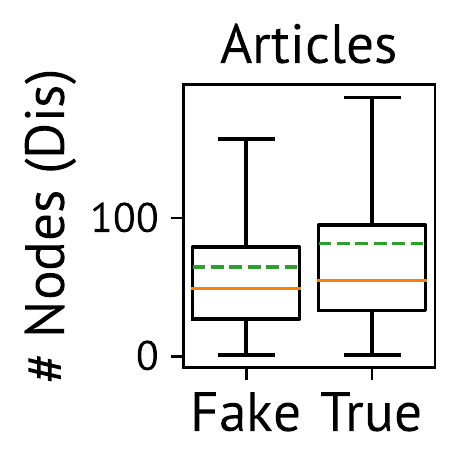}
        \includegraphics[width=0.1275\textwidth]{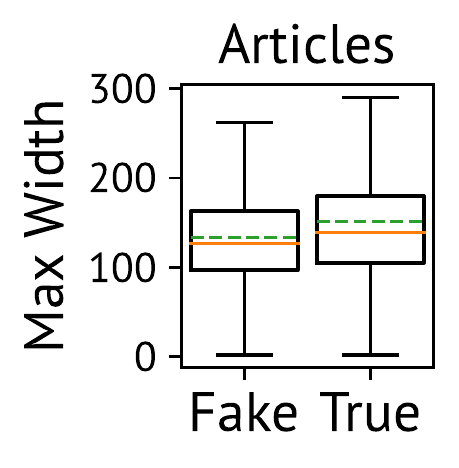}
        \includegraphics[width=0.12\textwidth]{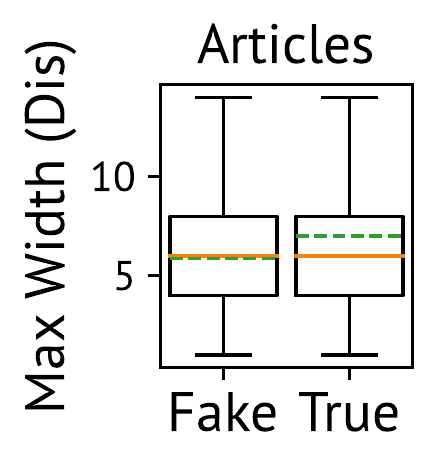}
    \end{minipage}
    }
    \caption{Hierarchical linguistic trees of fake and true news. Orange solid line: Median. Green dashed line: Mean.}
    \label{fig:hlt_of_fake_news_vs_true_news}
\end{figure*}

\subsection{Ablation Study}
\label{subsubsec:ablation_study}

We compare the proposed model, HERO, which contains hierarchical linguistic (syntax- and discourse-level) structures with its following variants.
\begin{itemize}
    \item \textit{HERO$\setminus$Dis}: It stands for the variant of HERO with only syntax-level structures. In this variant, the embedding of a news document is obtained by averaging its embeddings of EDUs.
    
    \item \textit{HERO$\setminus$Syn}: It stands for the variant of HERO with only discourse-level structures. In this variant, the embedding of each EDU of a news document is obtained by averaging its words.
    
    \item \textit{HERO$\setminus$(Syn+Dis)}: It stands for the variant of HERO with no structures, the embedding of a news document is directly obtained by averaging its word embeddings.
\end{itemize}

The results are visualized in Figure~\ref{fig:ablation_study}. We observe that with Recovery data, the proposed HERO outperforms HERO$\setminus$Dis by 1\% in AUC for unified HERO and by 3\% for level- and attribute-specific HEROs. It outperforms HERO$\setminus$Syn by 7--9\% and notably outperforms HERO$\setminus$Syn by above 30\% in AUC. With MM-COVID data, the proposed HERO performs similarly to HERO$\setminus$Dis since the statements presented in MM-COVID are short with two EDUs on average and hence have minimal discourse structures (i.e., syntax-level structures dominate hierarchical linguistic structures). Meanwhile, it outperforms HERO$\setminus$Syn and HERO$\setminus$(Syn+Dis) by more than 10\% in AUC.
Therefore, we conclude that the proposed HERO is better than its variants, demonstrating the importance of hierarchical linguistic structures.

\subsection{Characterizing Linguistic Style of Fake News}
\label{subsubsec:fake_news_patterns}

Fake news has been theoretically identified with a linguistic style distinguishable from the truth~\cite{undeutsch1967beurteilung}. This experiment aims to specify this different linguistic style of fake news. We compare the hierarchical linguistic trees generated by fake news and the truth, which we develop to represent the linguistic style of news documents systematically. The comparison is from the (i) children of parent nodes, (ii) attributes of nodes, and (iii) size, width, and depth of trees.

\paragraph{Children of Parent Nodes}
We compare fake news with real news in the average and the maximum number of children of parent nodes in hierarchical linguistic trees. Results that are statistically significant with a $p$-value $<0.001$ (using t-test, unless otherwise specified) are in Figure~\ref{subfig:children_of_parent_nodes}. We observe that the hierarchical linguistic trees of fake news have more child nodes for each parent node than true news on average. News here indicates long news articles in the Recovery dataset and short statements in the MM-COVID dataset. 

\paragraph{Attributes of Nodes}
Considering the nodes within a hierarchical linguistic tree can indicate the document (as the root), RRs, EDUs, POSs, and words (as the leaf nodes), we first compare fake news with the truth in the proportion of RRs, EDUs, POSs, and words, respectively. The reason for computing their proportions rather than the numbers is to eliminate the impact of the size of trees (discussed in the next paragraph). We observe that compared to true news, the hierarchical linguistic trees of fake news have a significantly smaller proportion of EDU nodes ($p$-value $<0.05$) and POS nodes indicating NNS (noun in the plural, $p$-value $<0.001$) but have a significantly larger proportion of nodes indicating specific POSs such as IN (preposition or subordinating conjunction), PP (prepositional phrase), and DT (determiner, $p$-value $<0.001$). We illustrate the results in Figure~\ref{subfig:node_attributes}.

\paragraph{Size, Width, Depth of Trees}  
We compare fake news with the truth in the size, maximum width, and depth of hierarchical linguistic trees. Since hierarchical linguistic trees contain two-level structures, we also compare fake and true news in the size, maximum width, and depth of discourse- and syntactic-level trees.

We observe that the syntactic-level tree of fake news is generally greater with more nodes, broader, and deeper than true news. In particular, the syntactic-level tree of fake news has more leaf nodes than true news, which reveals that fake news often has longer EDUs with more words than true news. The above conclusions hold for long news articles (using Recovery data) and short statements (with MM-COVID data) with a $p$-value $< 0.01$; news files in both datasets are rich in syntactic information. Figure~\ref{subfig:size_of_trees} (the upper ones) presents the details. Moreover, we observe that fake news articles generate smaller and narrower discourse-level trees that lead to smaller and narrower hierarchical linguistic trees than true news articles ($p$-value $< 0.01$, see the bottom figures in Figure~\ref{subfig:size_of_trees}).
We point out that the discourse structures of short statements are plain with two EDUs on average and hence have trivial impacts on the shape of the entire hierarchical linguistic structures. Lastly, we point out that comparing trees' maximum and average widths leads to the same conclusions. Comparing the longest (i.e., depth) and the average distance between the root and leaves also leads to the same conclusions.

\section{Conclusion}
\label{sec:conclusion}

We propose a psychology-informed neural network to predict fake news. The proposed neural network learns the linguistic style of news documents represented by hierarchical linguistic trees, which explicitly captures the writers' usage of words and the linguistically meaningful ways these words are structured as phrases, sentences, paragraphs, and, ultimately, documents. We conduct experiments on public real-world datasets. The results demonstrate the effectiveness of the proposed neural network, with 0.87--0.90 AUC scores, and the importance of the developed hierarchical linguistic tree. The proposed neural network can outperform the previous (recurrent, convolutional, graph, and self-attentive) neural networks and feature-engineering-based approach in predicting news--as long articles or short statements--as fake news or the truth. We observe from the data that the hierarchical linguistic trees of fake news can significantly differ from true news in the children of parent nodes, the attributes of nodes, and the size, width, and depth of the trees. In our future work, we aim to enhance the proposed model's performance with multimodal and social-context information.

\bibliographystyle{IEEEtran}
\bibliography{references}

\end{document}